\documentclass[10pt,twocolumn,letterpaper]{article}

\usepackage{wacv}    
\wacvfinalcopy
\usepackage{times}
\usepackage{epsfig}
\usepackage{graphicx}
\usepackage{amsmath}
\usepackage{amssymb}

\usepackage{booktabs,amsfonts,dcolumn}
\usepackage{float}
\usepackage{comment}
\usepackage{array}

\usepackage{epstopdf}
\usepackage{multirow}
\usepackage{xcolor}
\usepackage[switch]{lineno}  %
\usepackage{subfigure}
\usepackage{graphicx}
 
\usepackage{enumerate}
\usepackage{enumitem}

\usepackage{epstopdf}
\usepackage{multirow}
\usepackage{array}
\usepackage[linesnumbered,ruled]{algorithm2e}
\newcolumntype{L}[1]{>{\raggedright\let\newline\\\arraybackslash\hspace{0pt}}m{#1}}
\newcolumntype{C}[1]{>{\centering\let\newline\\\arraybackslash\hspace{0pt}}m{#1}}
\newcolumntype{R}[1]{>{\raggedleft\let\newline\\\arraybackslash\hspace{0pt}}m{#1}}
\usepackage[utf8]{inputenc}
\usepackage[english]{babel}
\usepackage{amsthm}
\usepackage[misc]{ifsym}

\usepackage[pagebackref,breaklinks,colorlinks]{hyperref}

\def\wacvPaperID{561}

\begin{document}

\title{ CoKe: Contrastive Learning for Robust Keypoint Detection }

\author{%
 Yutong Bai$^{1}$\thanks{Joint first authors}\;\;
 Angtian Wang$^{1}$\footnotemark[1]\;\;
 Adam Kortylewski$^{2,3}$\thanks{Joint senior authors}\;\;
 Alan Yuille$^{1}$\footnotemark[2]\\
  $^1$Johns Hopkins University\;\; $^2$University of Freiburg\;\; $^3$Max-Planck-Institute for Informatics\\
  \texttt{} \\
}

\maketitle

\begin{abstract}
In this paper, we introduce a contrastive learning framework for keypoint detection (CoKe). 
Keypoint detection differs from other visual tasks where contrastive learning has been applied because the input is a set of images in which multiple keypoints are annotated.
This requires the contrastive learning to be extended such that the keypoints are represented and detected independently, which enables the contrastive loss to make the keypoint features different from each other and from the background. 
Our approach has two benefits: It enables us to exploit contrastive learning for keypoint detection, and by detecting each keypoint independently the detection becomes more robust to occlusion compared to holistic methods, such as stacked hourglass networks, which attempt to detect all keypoints jointly. 
Our CoKe framework introduces several technical innovations. In particular, we introduce: (i) A clutter bank to represent non-keypoint features; (ii) a keypoint bank that stores prototypical representations of keypoints to approximate the contrastive loss between keypoints; and (iii) a cumulative moving average update to learn the keypoint prototypes while training the feature extractor.
Our experiments on a range of diverse datasets (PASCAL3D+, MPII, ObjectNet3D) show that our approach works as well, or better than, alternative methods for keypoint detection, even for human keypoints, for which the literature is vast. Moreover, we observe that CoKe is exceptionally robust to partial occlusion and previously unseen object poses.
\end{abstract}

\begin{figure}
    \centering
    \includegraphics[width=\linewidth]{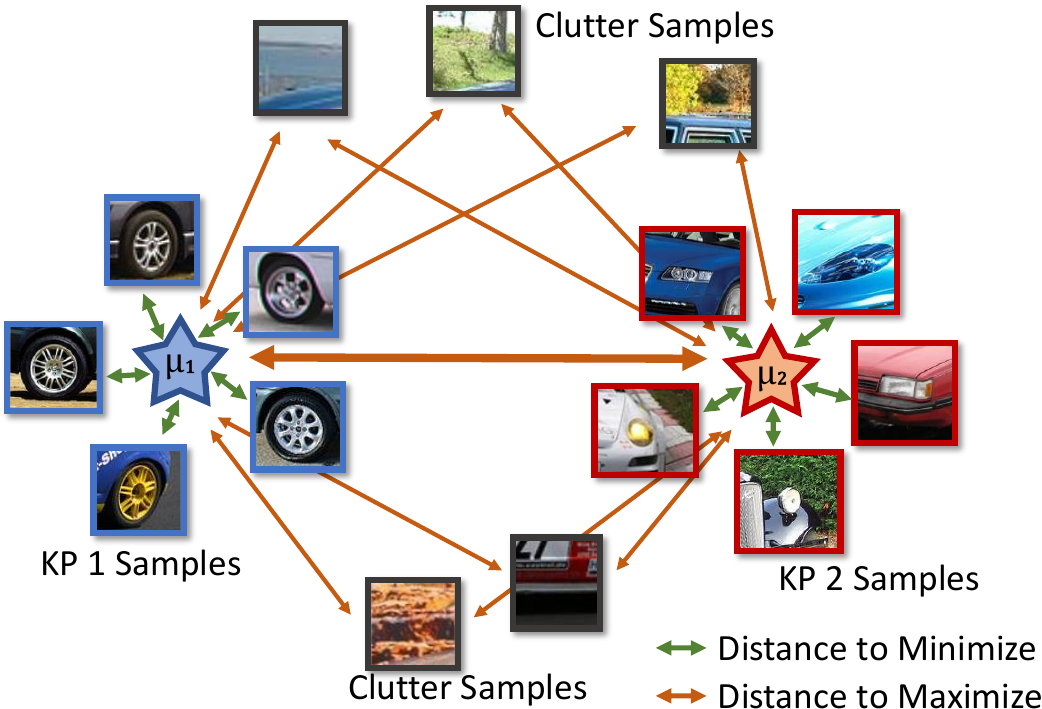}
    \caption{Intuition behind our approach.
    Image patches depict feature representations of two different keypoints (blue and red border) and background clutter (grey border). The star shapes illustrate the average representations $\mu_1$ and $\mu_2$ of the corresponding keypoint features. Our approach learns a representation space such that the following three distances are optimized: (1) The distance between features of the same keypoint is small, i.e. they cluster tightly around their mean. (2) The distance between the keypoint clusters is maximized. (3) The distance between the clutter features and the keypoint centers is maximized.}
    \label{fig:intro}
\end{figure}

\section{Introduction}
\label{sec:introduction}

Semantic keypoints, such as the joints of a human body, provide concise abstractions of visual objects in terms of their shape and pose. Accurate keypoint detections are of central importance for many visual understanding tasks, including viewpoint estimation \cite{pavlakos20176}, human pose estimation \cite{cao2017realtime}, action recognition \cite{messing2009activity}, feature matching \cite{long2014convnets}, image classification \cite{zhang2014part}, and 3D reconstruction \cite{kanazawa2018learning}. There are many diverse approaches for keypoint detection. Common approaches include the application of a regression loss \cite{newell2016stacked, ke2018multi}, a classification loss \cite{he2017mask}, or combinations of either of those with a 3D geometric model of the object \cite{zhou2018starmap}. 
In recent years, work in contrastive learning has led to major advances in representation learning \cite{chen2020simple, he2019momentum, misra2020self} demonstrating benefits over classical losses, such as cross-entropy, e.g. in terms of robustness and data efficiency \cite{khosla2020supervised}.
However, most works on contrastive learning in computer vision focus on the task of image classification, and it remains unclear how contrastive learning can be applied for keypoint detection.

In this paper, we introduce a contrastive learning framework for keypoint detection (CoKe). 
Keypoint detection differs from other visual tasks where contrastive learning has been applied, such as face recognition \cite{schroff2015facenet} or unsupervised learning \cite{he2019momentum},
because the input is a set of images in which multiple keypoints are annotated.
To enable the contrastive learning of keypoint detectors, we need to represented and detected keypoints independently, such that a contrastive loss can make the keypoint features different from each other and from the background.
This is very different from current popular approaches to keypoint detection, such as stacked hourglass networks \cite{stack}, which attempt to detect all keypoints jointly. 
Our approach has two benefits: It enables us to exploit contrastive learning for keypoint detection, and by detecting each keypoint independently the detection can become more robust to occlusion compared to holistic methods (as shown in our experiments).

Our CoKe framework introduces several technical innovations. Specifically, we found that the contrastive learning of keypoint representations requires the optimization of three types of distances in the feature space (Figure \ref{fig:intro}):

(1) The distance between features of the same keypoint should be small. But the computational cost to calculate the distance between all features of the same keypoint is quadratic in their number. 
We instead reduce the computational cost to be linear by introducing an average prototypical representation for each keypoint (stars in Figure \ref{fig:intro}). 

(2) The distance between features of different keypoints should be large. The number of distance comparisons between features of different keypoints is combinatorial in the number of keypoints and training images. 
To manage this computational burden, we introduce a \textit{keypoint bank} which stores the prototypical representations of all keypoints and allows for an efficient computation of the distance between the prototypes of different keypoints (Figure \ref{fig:intro} bold orange arrow).

(3) The distance between keypoint features and features from the background clutter (Figure \ref{fig:intro} grey squares) should be large to reduce false-positive detections. However, most of the features in an image are clutter features and it is not feasible to compute the distance to all of them. Therefore, we introduce a \textit{clutter bank} that keeps track of clutter features that are spatially close to keypoint features and hence are most difficult to be distinguished from.

The proposed approximations enable an efficient contrastive learning of keypoint detectors.
We evaluate CoKe on several datasets including PASCAL 3D+, MPII, and ObjectNet3D. We observe that CoKe performs on par, and often even better, compared to SOTA related work (Stacked Hourglass Networks, MSS-Net~\cite{ke2018multi}) and to approaches that use additional supervision in terms of detailed 3D object geometries (StarMap~\cite{zhou2018starmap}).
These results are remarkable as CoKe works well on all these datasets while, for example, the best results on MPII are often achieved by architectures that are specialized for human keypoint detection. 
We also observe that when compared to related work, CoKe is exceptionally robust to partial occlusion and unseen object poses.
Our main contributions are:
 \begin{enumerate}
     \item We introduce a contrastive learning framework for keypoint detection. 
    \item CoKe performs very well on various keypoint detection datasets for rigid and articulated objects. 
    \item CoKe achieves exceptional robustness to partial occlusion and previously unseen object poses.
\end{enumerate}

\section{Related Work}
\label{sec:related_work}
\textbf{Keypoint Detection.}
Keypoint detection, is a widely studied problem in computer vision. Popular applications are, e.g., the detection of human joints \cite{cao2017realtime,newell2016stacked,tompson2015efficient,toshev2014deeppose} or distinct locations on rigid objects \cite{wu2016single, tulsiani2015viewpoints,pavlakos20176,zhou2018starmap}.
Early approaches relied on local descriptors  \cite{gourier2004facial, schmidt2016self, yi2016lift, choy2016universal, florence2018dense} that are distinctive and invariant \cite{lowe2004distinctive}.
While approaches using local descriptors have proven to be robust to occlusion and background clutter, they were outperformed by deep learning approaches that were trained end-to-end \cite{newell2016stacked}. 
Toshev et al. \cite{toshev2014deeppose} first trained a deep neural network for 2D human pose regression and  Li et al. \cite{li20143d} extended this approach to 3D. 
Starting from the work of Tompson et al. \cite{tompson2015efficient}, regression-based approaches to keypoint detection became very popular. They perform keypoint detection by regressing a heatmap representation. These approaches achieve a particularly good performance at detecting the joints of both articulated and rigid objects \cite{newell2016stacked}, because they can implicitly leverage the structural information between keypoint to resolve locally ambiguous keypoint detections.   
Tulsiani et al. \cite{tulsiani2015viewpoints}, proposed to integrate the structural information between keypoints explicitly by integrating 2D and 3D models, which inspired a number of follow-up works, in particular for rigid objects \cite{zhou2018starmap,tulsiani2015viewpoints,pavlakos20176}. 

\textbf{Supervised Contrastive Learning.}
Contrastive learning, originates from Metric Learning \cite{chechik2009large,weinberger2009distance, rippel2015metric} and involves the learning of a representation space by optimizing the similarities of sample pairs in this space. 
Intuitively, supervised contrastive learning aims to reduce the distance of feature representations of the same class, while increasing the distance between samples from different classes.
Popular examples use pairs of samples for loss computation \cite{hadsell2006dimensionality}, triplets \cite{schroff2015facenet}, or N-Pair tuples \cite{sohn2016improved}. 

Recently, contrastive learning has attracted attention from the research community in  self-supervised learning \cite{chen2020simple, he2019momentum, wu2018unsupervised, misra2020self, henaff2019data}. The main difference in the self-supervised learning setting is that positive examples are usually generated using data augmentations\cite{cubuk2018autoaugment} or co-occurrence \cite{hjelm2018learning, tian2019contrastive} of a query sample, whereas negative examples are chosen as other images in the same mini-batch.

While most of the supervised contrastive learning \cite{khosla2020supervised} focuses on learning a holistic representation of the complete image, in this paper, we target a more fine-grained task - keypoint detection. 
Keypoints are localized image patterns and therefore require the learning of local feature embeddings. 
The main challenge is that local image patterns can be highly ambiguous (e.g. the front and back tire of a car) and therefore require a contrastive learning framework that can learn to disambiguate local representations, while at the same time being able to learn a distinct representation that can be localized accurately. 

\begin{figure*}
\centering
\includegraphics[width=0.92\linewidth]{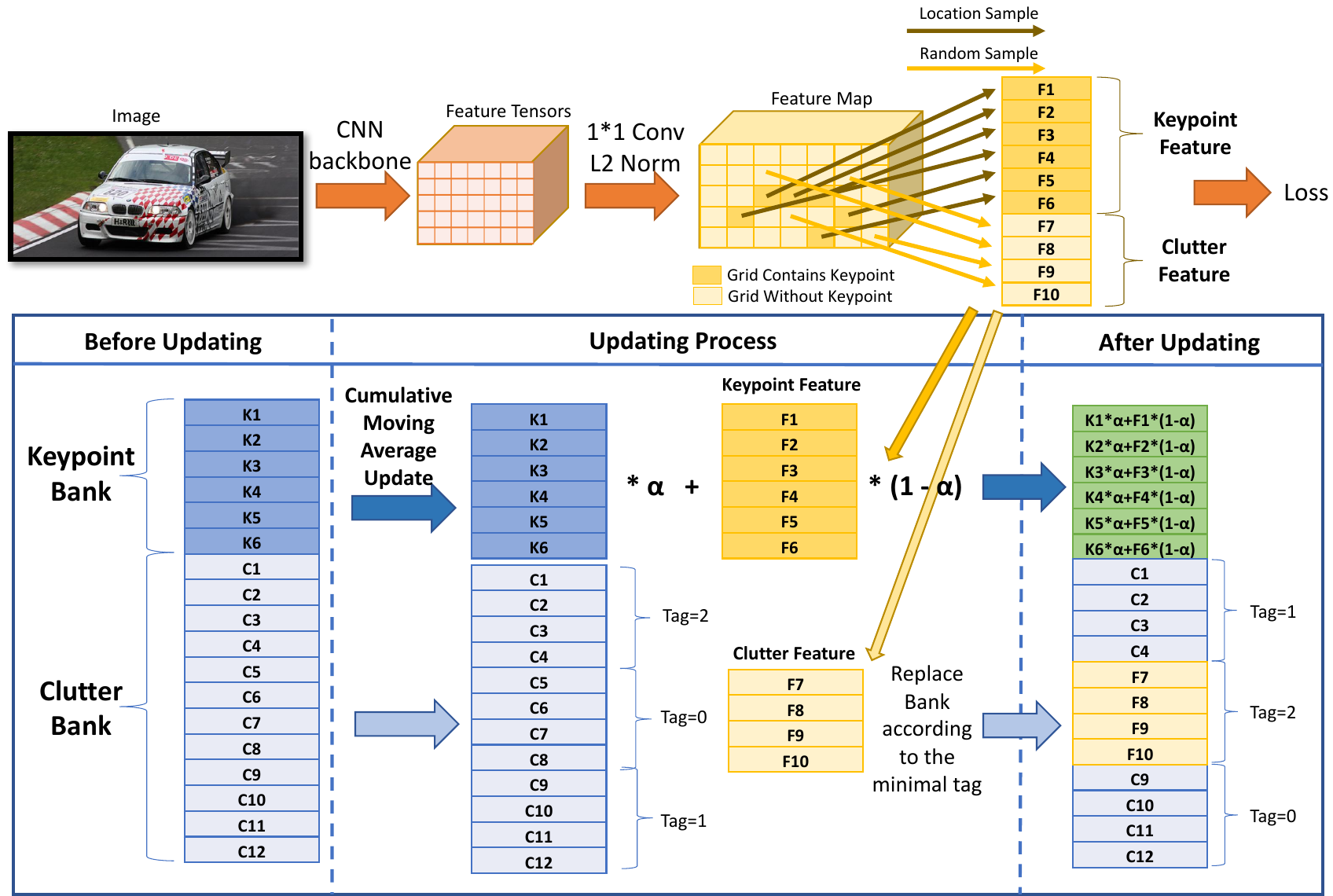}
\caption{Illustration the Keypoint and Clutter Bank update process. First, a feature map for the input image is extracted. After a dimensionality reduction and $L_{2}$ normalization, we retrieve the keypoint features (F1-F6) and randomly selected clutter features (F7-F10).
The Keypoint Bank is updated using a cumulative moving average update. The Clutter Bank is updated by replacing the oldest features in the Clutter Bank with (F7-F10) based on the time tag. }
\label{fig:update_bank} 
\end{figure*}

\section{CoKe: Contrastive Keypoint Learning}
In this section, we present our framework for contrastive keypoint learning, then discuss the intuition underlying our approach and the training pipeline and how to perform keypoint detection during inference.

\subsection{Training CoKe}
We use $\Phi$ to denote the feature extractor. Given an input image $\mathbf{I}^i$, it computes the feature map $ \Phi (\mathbf{I}^i) = \mathbf{F}^i\in \mathbb{R}^{H \times W \times D}$, where $i$ is the image index in the training data $\{\mathbf{I}^i|i\in\{1,\dots,N\}\}$.
Using the keypoint annotation we retrieve the corresponding keypoint features $\mathbf{f}_k^i \in \mathbb{R}^D$ for the keypoints $k \in \mathbb{K}$ from the feature map $\mathbf{F}^i$.
Similarly, we can (randomly) select a non-keypoint location as clutter point and retrieve a set of clutter features $\{\mathbf{f}_c^i\in \mathbb{R}^D | c\in\{1,\dots,C\}\}$.
We define the distance between two features as $d(\cdot, \cdot)$. 
During training we learn a feature extractor that optimizes the following distances in the feature space:

(1) One objective for the feature extractor during training is to \textbf{minimize the distance between features of the same keypoint} (i.e. the intra-keypoint distance) across all training images, by minimizing the objective:
\begin{equation}
    D_{intra}(\mathbf{f}_{k}^i) = \sum_{j = 1}^N d(\mathbf{f}_k^i, \mathbf{f}_k^j)\ \approx d(\mathbf{f}_k^i, \mathbf{\mu}_k).
\end{equation} 
However, as we described in the introduction it is unpractical to compute the distance of a keypoint's feature vector from one image $\mathbf{f}_k^i$ to the corresponding vectors in all other training images $j \in\{1,\dots,N\}$.
To resolve this computational problem, we define a prototypical keypoint feature $\mathbf{\mu}_k$, that represents the average feature of keypoint $k$. 
Instead of computing the full objective, we approximate it as simply the distance to the corresponding average representation $d(\mathbf{f}_k^i, \mathbf{\mu}_k)$.
We store the prototypical features of all keypoints $\{\mu_k\}$ in a \textit{keypoint bank} and update them during training.

(2) The second objective for the feature extractor is to \textbf{maximize the distance between features of different keypoints} (i.e. the inter-keypoint distance). 
This requires computing the distance between the feature representations of one particular keypoint $k$ and all other keypoints $k^{'}$ over all training images:
\begin{equation}
D_{inter}(\mathbf{f}_{k}^i) = \sum_{k' \in K\setminus \{k\}} \sum_{j = 1}^N  d(\mathbf{f}_{k}^i, \mathbf{f}_{k'}^j) \approx \sum_{k' \in K\setminus \{k\}} d(\mathbf{f}_{k}^i, \mathbf{\mu}_{k'}).
\end{equation}
We approximate this objective by instead computing the distance to the prototypes of the respective keypoint features from the keypoint bank. 

(3) The third objective for the feature extractor is to \textbf{maximize the distance between the keypoint features and all clutter features}. 
In the best case, this involves computing the distance between a
keypoint feature $\mathbf{f}_k^i$ to every clutter feature in all training images.
To avoid the computation of this large amount of distances, we instead approximate this objective by storing a subset of all clutter features $\{\mathbf{\theta}_c, c \in \mathbb{C}\}$ in a \textit{clutter bank}. Which allows us to approximate the full objective with 
\begin{equation}
    D_{clutter}(\mathbf{f}_k^i) \approx \sum_{c \in \mathbb{C}} d(\mathbf{f}_{k}^i, \mathbf{\theta}_c).
\end{equation}

These three approximations make it feasible to optimize the overall objective with a feasible computational load. As the parameters of the feature extractor will change during learning, the clutter features $\mathbf{f}_c$ in the clutter bank as well as the prototypes $\mathbf{\mu}_k$ need to be updated.
To achieve this, we follow an EM-type optimization process. First, we initialize the clutter bank by randomly sampling clutter features from the training data, and initialize the prototypes by computing the average feature for every keypoint across the training data $\mu_k=\frac{\sum_{i=1}^N \mathbf{f}_k^i}{N}$.
Using these initial estimates, we can compute the overall objective and train the feature extractor.
While training the feature extractor, we update the clutter and keypoint bank. We perform those updates in an alternating manner.
\subsubsection{Keypoint and Clutter Bank Update}
\label{sec:band_update}
Figure \ref{fig:update_bank}, illustrates the process of updating the keypoint prototypes and the clutter bank during training.

\textbf{Keypoint Bank Update.}  Computing the prototypical keypoint features $\mathbf{\mu}_k$ while learning the feature extractor is challenging, because we want to avoid to re-compute the prototypes over all training images $\mu_k=\frac{\sum_{i=1}^N \mathbf{f}_k^i}{N}$ after every gradient step. 
Instead, we approximate the sample mean via a cumulative moving average. %
Specifically, we update $\mathbf{\mu}_k$ with a training batch of size $m$ using:
\begin{equation}
    \mathbf{\mu}_k \leftarrow \mathbf{\mu}_k * \alpha + \frac{\sum_{i=0}^m \mathbf{f}_k^i}{m} * (1 - \alpha).
\end{equation}

\textbf{Clutter Bank Update.} The clutter bank contains a limited number $C$ of clutter features $\{\mathbf{\theta}_c, c \in \mathbb{C}\}$. In practice the size of the clutter bank depends on the availability of GPU memory and we observe that the larger the bank, the better the training performance (see experiments section). We update the clutter bank during training by replacing the oldest clutter features with newly extracted features from the current training batch based on a time tag that indicates how long features were stored in the bank. 

\subsubsection{Feature Extractor Training}
\label{sec:ext_update}

When training the feature extractor with SGD, we freeze the keypoint bank and clutter bank in every gradient step and use them to compute the gradient update on the weights of the feature extractor. 
To compute the distance between two feature vectors, we use the L2 distance:
\begin{equation}\label{eq:cos}
    d(\mathbf{f}_a,\mathbf{f}_b)=(\mathbf{f}_a - \mathbf{f}_b)^2 = 2 * (1 - \mathbf{f}_a \cdot \mathbf{f}_b).
\end{equation}
The last step in above equation leverages that all features in our model are L2 normalized.
From Eq. \ref{eq:cos} observe that we can minimize $D_{intra}(\mathbf{f}_k^i)$ by maximizing $\mathbf{f}_k^i \cdot \mathbf{\mu}_k$. 
Similarly, we can maximize $D_{inter}(\mathbf{f}_k^i)$ and $D_{clutter}(\mathbf{f}_k^i)$ by minimizing $\{\mathbf{f}_{k}^i \cdot \mathbf{\mu}_{k'} | \forall k'\in K\setminus \{k\}\}$ and $\{\mathbf{f}_{k}^i \cdot \mathbf{\theta}_c\ |\forall c \in \mathbb{C}\}$ respectively.
To optimize those terms simultaneously, we use a non-parametric softmax as our loss function. Thus, the loss for each keypoint feature is calculated as:
\begin{equation}
    \mathcal{L}(\mathbf{f}_{k}^i,\{\mu_k\},\{\mathbf{\theta}_c\}) = \frac{e^{\mathbf{f}_{k}^i \cdot \mathbf{\mu}_{k}}}{\sum_{k^{'} \in K} e^{\mathbf{f}_{k}^i \cdot \mathbf{\mu}_{k^{'}}} + \sum_{c^{'} \in C} e^{\mathbf{f}_{k}^i \cdot \mathbf{\theta}_{c^{'}}}},
\label{equ:model:keypoint_loss}
\end{equation}
where $\{\mu_k\}$ is the keypoint bank and $\{\theta_c\}$ is the clutter bank.

\textbf{Clutter Sampling Loss.} 
A technical problem is that the features in the clutter bank are copied from a large number of training images, and therefore it is not practical to calculate gradient directly w.r.t. 
$\{\theta_c\}$, and therefore the feature extractor is not optimized w.r.t. the clutter features. This technical limitation makes the optimization w.r.t. the clutter distance $D_{clutter}(\mathbf{f}_k^i)$ converge slowly, especial when the clutter bank is large. 
To make the training more efficient, we propose a clutter sampling loss, which uses the sampled clutter features $\mathbf{f}_{c}^i$ from the current training batch and directly maximizes the clutter to keypoint distance:
\begin{equation}
    \mathcal{L}(\mathbf{f}_{c}^i, \{\mu_k\}) = \sum_{k^{'} \in K} \mathbf{f}_{c}^i \cdot \mathbf{\mu}_{k^{'}}.
\label{equ:model:clutter_loss}
\end{equation}
Thus, the final loss for training the feature extractor becomes:
\begin{equation}
    \hspace{-.1cm}\mathcal{L}(\mathbf{F}^i{,}\{\mu_k\}{,}\{\mathbf{\theta}_c\}) {=}\hspace{-.1cm} \sum_{k \in \mathbb{K}} \mathcal{L}(\mathbf{f}_{k}^i,\{\mu_k\},\{\mathbf{\theta}_c\}) {+}\hspace{-.1cm}\sum_{c \in \mathbb{C}} \mathcal{L}(\mathbf{f}_{c}^i, \{\mu_k\}).\hspace{-.4cm}
\label{equ:model:loss}
\end{equation}

\begin{figure*}
\centering
\includegraphics[width=\linewidth]{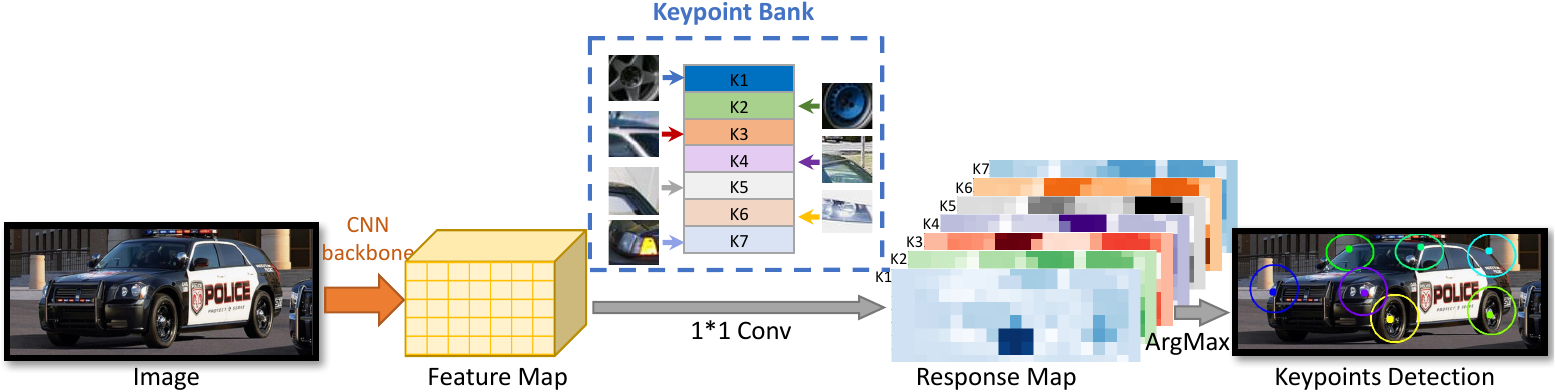}
\caption{Keypoint detection with CoKe. We first extract feature representation at different positions of input image. Each keypoint in the Keypoint Bank has an individual representation used as a convolution kernel to compute a response map. The location of maximum response is used as prediction result. Colored boxes show ground truth and dots the prediction result.}
\label{fig:pipeline}
\end{figure*}

\subsection{Inference with the CoKe Model}
\label{sec:coke:pipeline}
In the following, we describe the detection process for keypoints of a single category, but note that this can be trivially extended to multiple categories. Figure \ref{fig:pipeline} illustrates the inference process with the CoKe.
To locate the predicted the keypoint position $p_k$ for keypoint $k$ on the feature map of a test image $\mathbf{I}^i$, we apply the following steps:

    Extract a feature map $\Phi (\mathbf{I}^i) = \mathbf{F}^i$ using the trained feature extractor $\Phi$.
    Use the prototypes ${\mathbf{\mu}_k, k \in \mathbb{K}}$ to compute the per pixel feature distance $d(\mathbf{f}, \mathbf{\mu}_k)$ for all feature vectors $\mathbf{f} \in \mathbf{F^i}$
    and store the detection scores in the output $\mathbf{S} \in \mathbb{R}^{H \times W \times K}$.
    For each keypoint, select the position $p$ with the highest detection score in $\mathbf{S}$.
    Project $p$ back to the original image coordinates.

\section{Experiments}
\label{sec:experiment}
In this section, we experimentally evaluate CoKe and compare it to related works. 
We first describe the experimental setup, compare CoKe to related work on several diverse datasets, and study their robustness to partial occlusion. Then discuss qualitative results with ablation studies.

\subsection{Experimental Setup}
\label{sec:experiment:setup}
\textbf{Evaluation Protocol.} Evaluation is done using the standard Percentage of Correct Keypoints (PCK) metric which reports the percentage of detections that fall within a normalized distance of the ground truth. we use PCK=0.1 for PASCAL3D+, ObjectNet3D and OccludedPASCAL3D+ datasets following the standard experimental protocol. 
For MPII, we use PCKh=0.5 as the evaluation metric. Distance is normalized by a fraction of the head size (PCKh). We follow the common protocol of evaluating each object category by computing the average accuracy on the visible keypoints over all the test images. 

\noindent \textbf{Training Setup.} We  use  the  standard  train-val-test  split  for  all the datasets. We use a batch size of $64$ for training. For each image, we randomly select a group of $20$ clutter points. The full clutter bank consists of $1024$ such groups. We choose the clutter features to be within two pixels distance to the keypoint annotation in the feature map. We use the non-parametric  softmax \cite{wu2018unsupervised} to calculate the similarity between features and banks. The temperature parameter\cite{hinton2015distilling} that controls the concentration level of the distribution is set to $\tau=0.7$. 
\begin{figure*}[!t]
\centering 
\label{FalsePositives}
\includegraphics[width=0.24\textwidth]{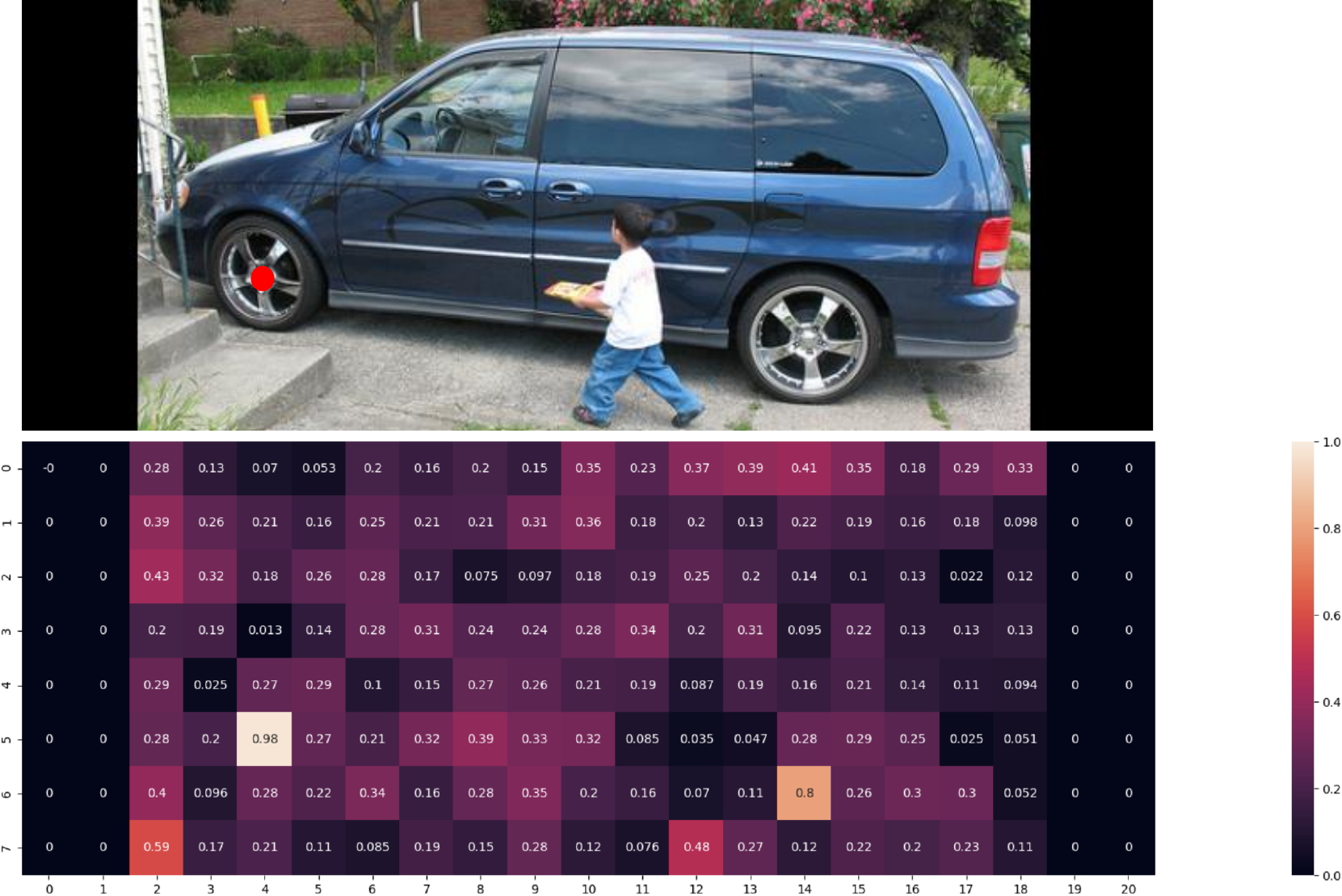}
\includegraphics[width=0.24\textwidth]{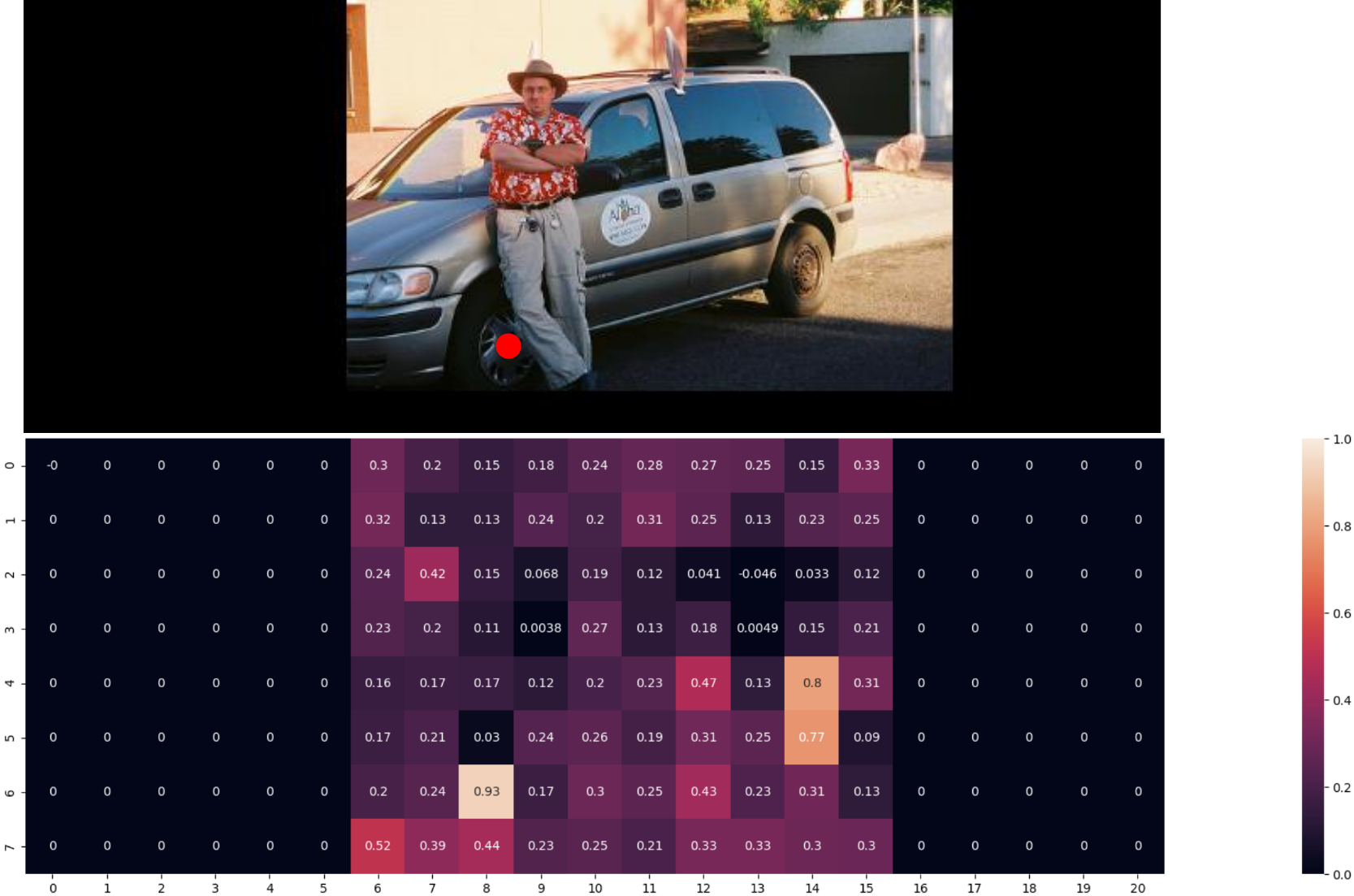}
\includegraphics[width=0.24\textwidth]{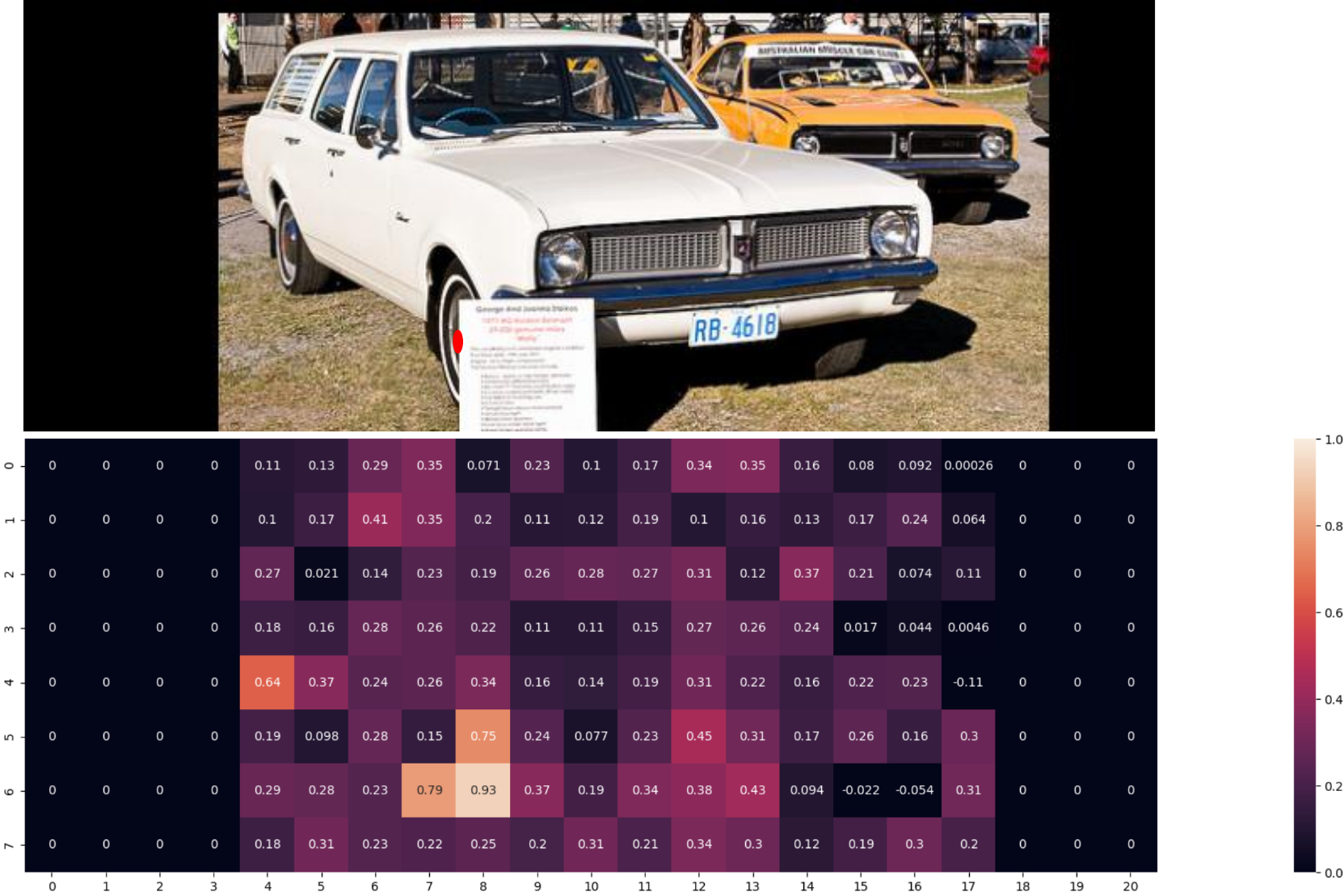}
\includegraphics[width=0.24\textwidth]{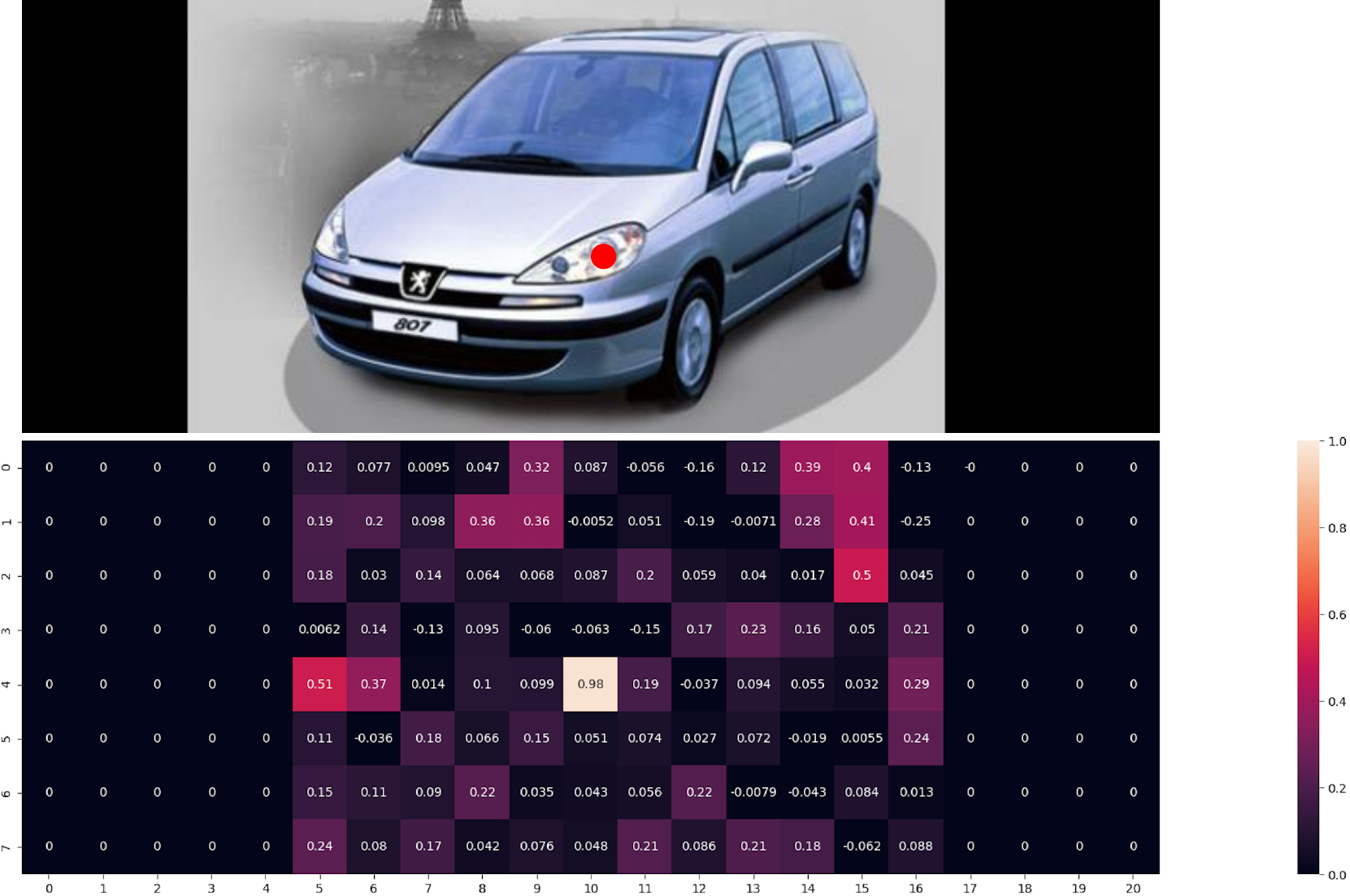}
\includegraphics[width=0.24\textwidth]{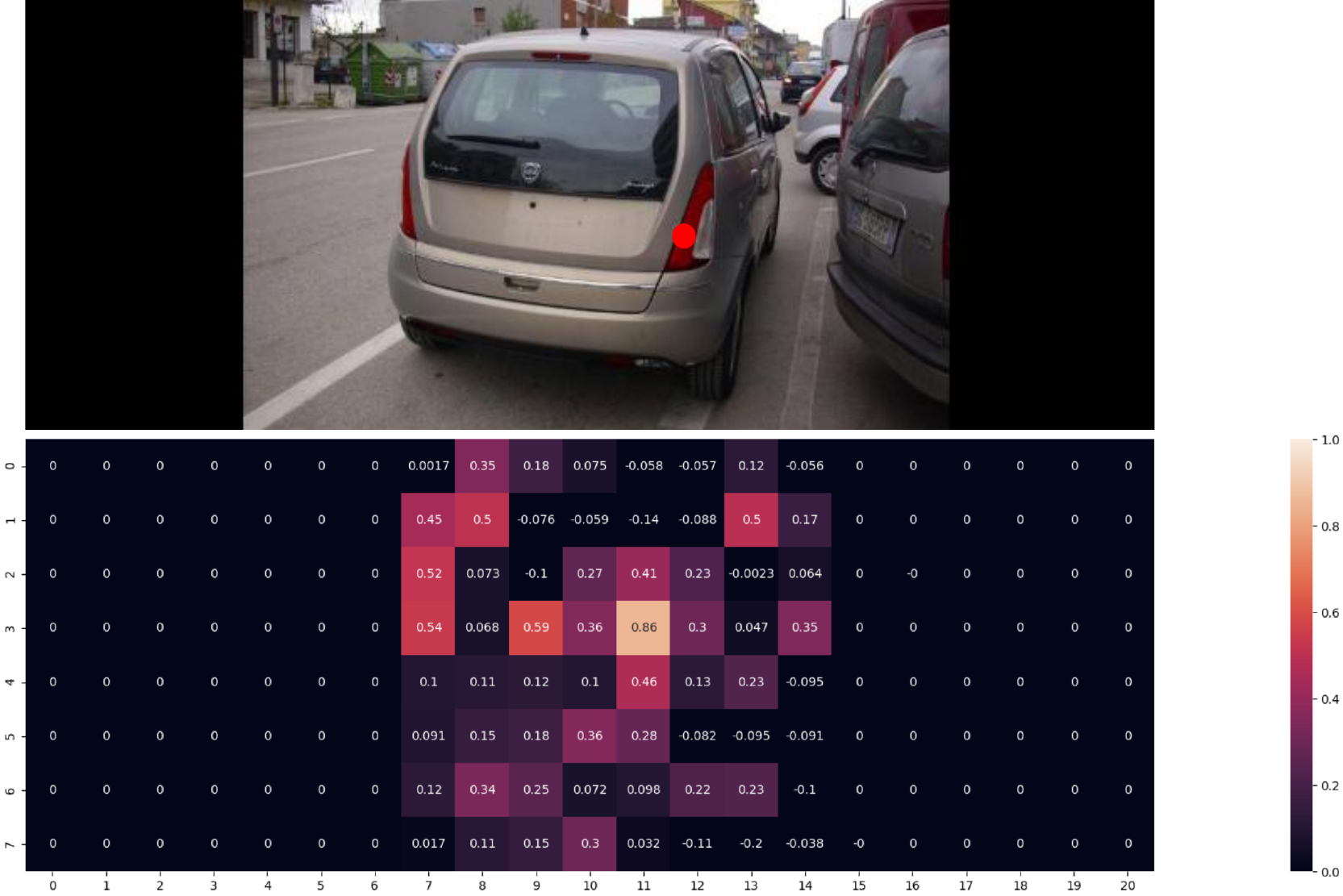}
\includegraphics[width=0.24\textwidth]{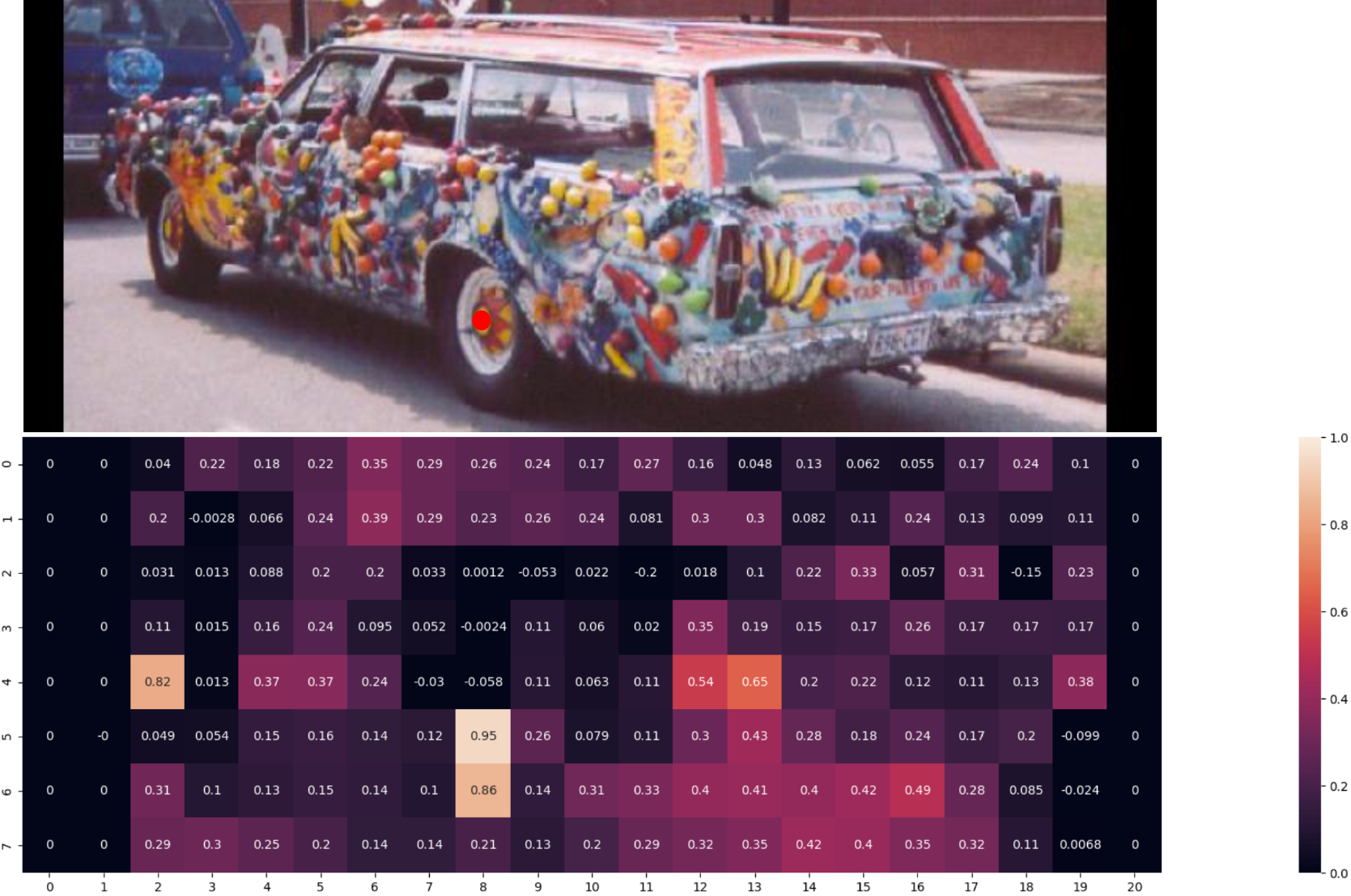}
\includegraphics[width=0.24\textwidth]{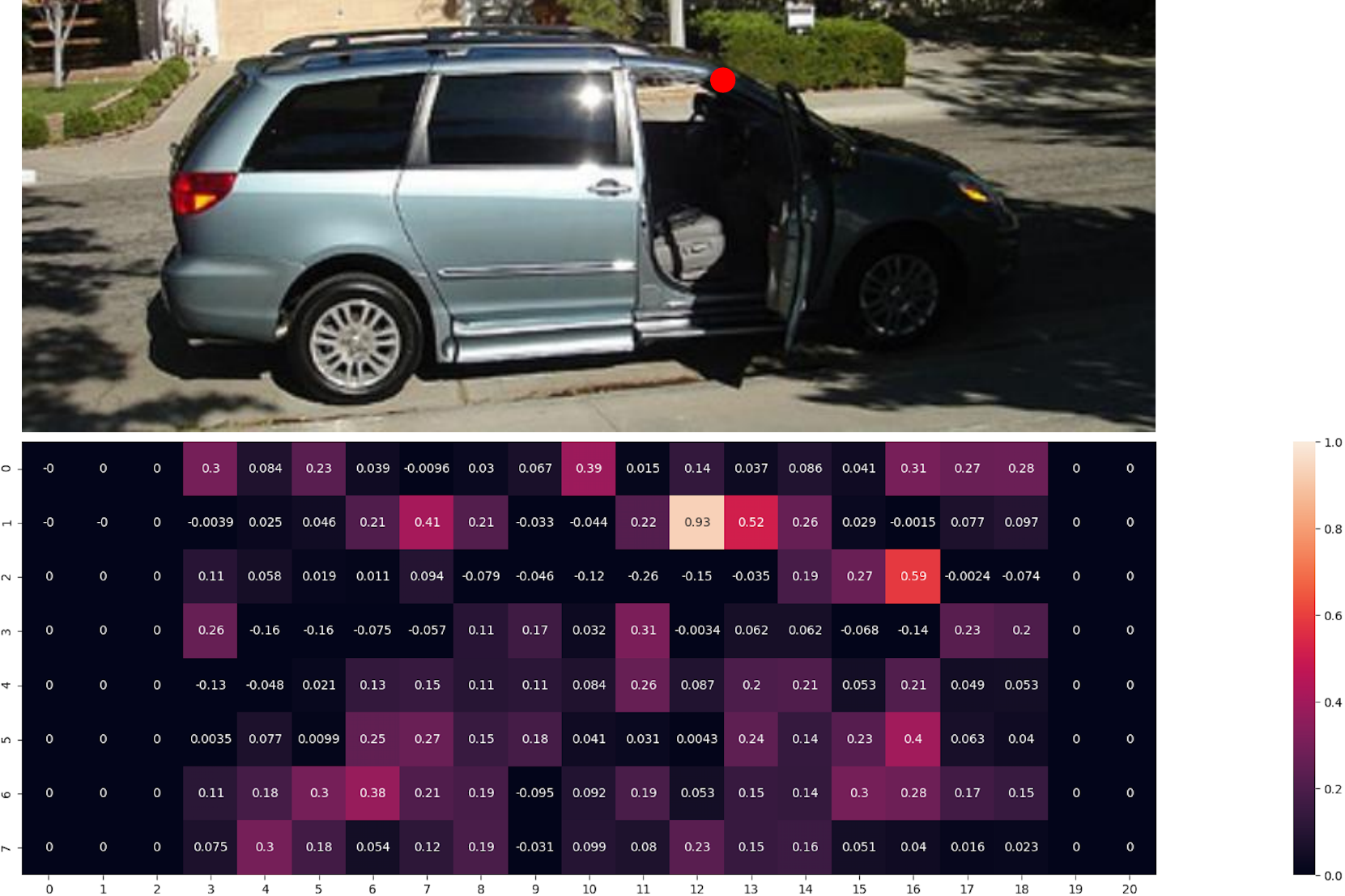}
\includegraphics[width=0.24\textwidth]{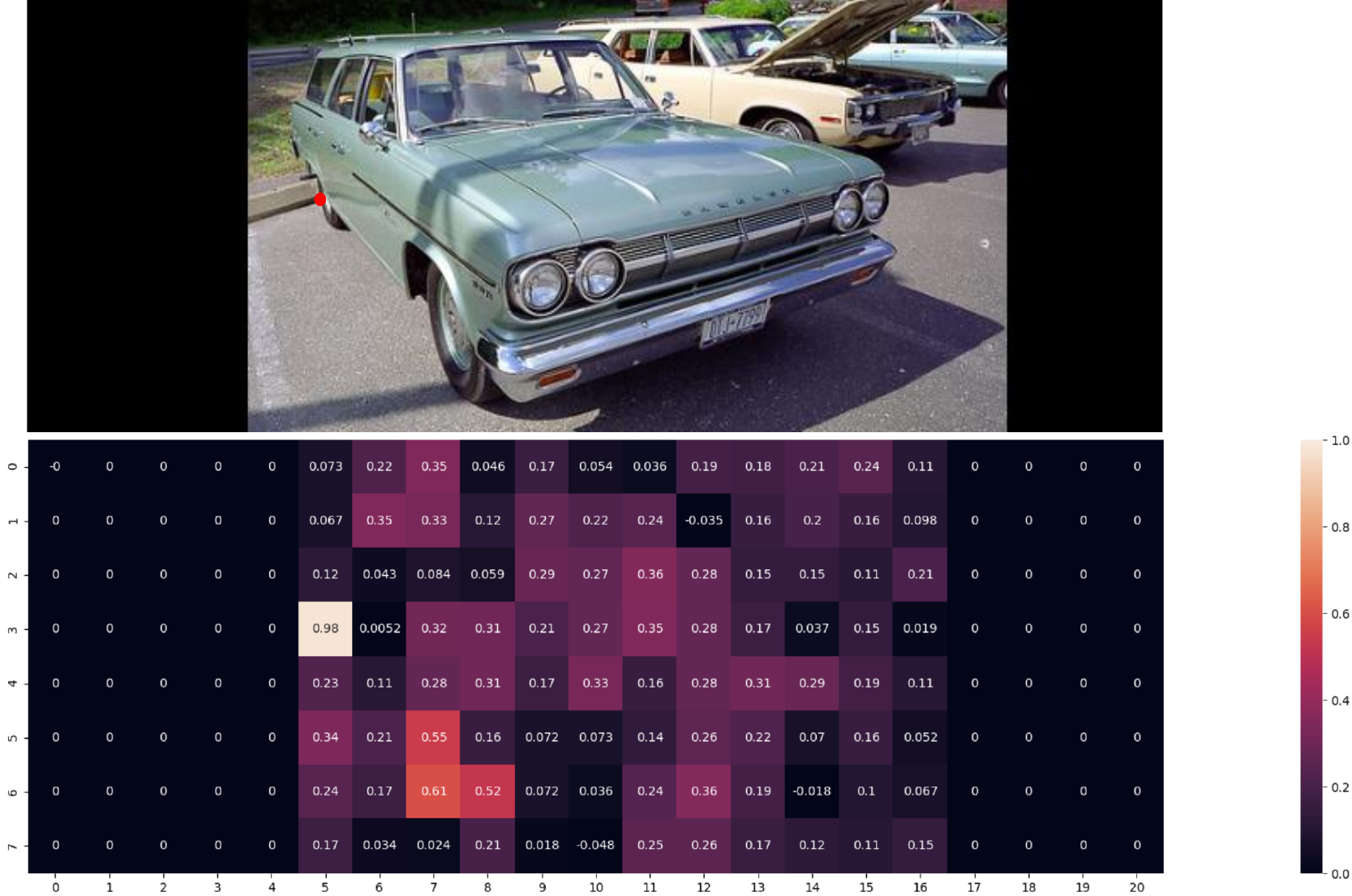}
 \caption{Eight examples of CoKe-Res50's representation visualization.  For each sub-figure, top is the original image with keypoint annotation, labeled with red dot. Bottom is the response map, predicted by CoKe-Res50. It is worth noting that how all keypoints are detected accurately despite the difficulty from false positives, occlusions, rare viewpoints, different domain, irregular appearance and irregular state and rare viewpoints. }
\label{fig:featuremap}
\end{figure*}

\textbf{PASCAL3D+ Dataset.} 
The dataset contains 12 man-made object categories with totally training 11045 images and 10812 evaluation images. Different from previous works \cite{zhou2018starmap,tulsiani2015viewpoints}, we use all images for evaluation, including occluded and truncated ones.

\textbf{MPII Dataset.} 
MPII Human Pose \cite{andriluka14cvpr} 
consists of images taken from a wide range of human activities with a challenging array of articulated poses. The keypoint visibility is annotated, enabling us to report numbers for the full dataset as well as partially occluded humans.

\textbf{ObjectNet3D Dataset.} ObjectNet3D consists of common daily life objects and is notably more difficult compared to PASCAL3D+ as it contains more rare viewpoints, shapes and truncated objects with occlusion. We test on nine categories, chosen based on their high annotation accuracy, as some categories in ObjectNet3D have low quality annotations due to the complexity of this dataset.

\textbf{OccludedPASCAL3D+ Dataset.} While it is important to evaluate algorithms
on real images of partially occluded objects, simulating occlusion enables us to quantify the effects of partial occlusion more accurately. 
We use an analogous dataset with artificial occlusion for keypoint detection proposed in \cite{wang2020compositional} for object detection. It contains all 12 classes of the PASCAL3D+ dataset at various levels of occlusion. 
The dataset has a total of 3 occlusion levels, with Lv.1: 20-40\%, Lv.2: 40-60\% and Lv.3: 60-80\% of the object area being occluded.

\subsection{Performance on Various Datasets}
\label{ref:exp:main}
\textbf{PASCAL3D+ and OccludedPASCAL3D+.} 
Table \ref{tab:experiment:p3d+} shows the keypoint detection results on the PASCAL3D+ dataset for CoKe models learned from three different backbones: ResNet-50 \cite{he2016deep}, Stacked-Hourglass-Network and Res-UNet \cite{zhang2018road}. 
We also show the performance of StarMap \cite{zhou2018starmap} as reported in the original paper. Note that StarMap uses 3D models as additional supervision to jointly reason about the relative position of the keypoints.
The performance of CoKe with all backbones is constantly high. 
The highest performance is achieved with the most recently developed architecture Res-UNet. 
When compared to the original Stacked-Hourglass-Network trained with a regression loss (SHG) we can clearly observe a large gain in performance. Most notably, the performance difference is very prominent for strong occlusion.
We believe CoKe is more robust to occlusion, because of two reasons: 1) It is actively optimized to discriminate between keypoint features and clutter features based on their local representations only, which might help to reduce the effective receptive field size and hence reduces the negative effects of occlusions. 2) CoKe stores the clutter representation from a large number of images, and hence can better learn to distinguish keypoints from clutter. 
Overall, our results clearly highlight that CoKe is very competitive with related work, while being highly robust to partial occlusion. 

In Table \ref{tab:experiment:unseenview}, we demonstrate an additional robustness to \textbf{unseen object poses}.
We divide the azimuth pose in the \emph{car} category of Pascal3D+ into 4 bins, front, back, left and right. We train CoKe on the front and back subsets, and test on seen (7305 images) and unseen (3507 images) poses separately. Table \ref{tab:experiment:unseenview} shows that CoKe is much more robust to unseen poses compared to SHG.

\begin{table}[h]

    \centering
    \begin{tabular}{l|ccccc}
      \hline
      \multicolumn{6}{c}{\textbf{PASCAL3D+}} \\
      \hline
        Occlusion Level & Lv.0 & Lv.1 & Lv.2 & Lv.3 & Avg \\
      \hline
      SHG & 68.0 & 46.5 & 43.2 & 39.9 &49.4\\
      MSS-Net & 68.9 & 46.6 & 42.9 & 39.6 & 49.5\\
      StarMap& 78.6 & - & - & - & -\\
      \hline
      CoKe-Res50 & 77.0 & 67.6& \textbf{59.9} & 53.4 &64.4\\ 
      CoKe-SHG & 78.3 & 66.3& 58.4 & 52.3&63.8\\ 
      CoKe-ResUnet & \textbf{80.3} & \textbf{68.5} & 59.1 & \textbf{54.0}&\textbf{65.5}\\ 
      \hline
    \end{tabular}
    \caption{Keypoint detection results on PASCAL3D+ under different levels of partial occlusion (Lv.0:0\%,Lv.1:20-40\%,Lv.2:40-60\%,Lv.3:60-80\% of objects are occluded, L0 is the original dataset). CoKe outperforms baseline models and is highly robust to partial occlusion. }
    \label{tab:experiment:p3d+}
\end{table}

\begin{table}[h]
\centering

\begin{tabular}{l|cc}
 \hline
 \multicolumn{3}{c}{\textbf{PASCAL3D+}} \\
 \hline
& Seen-Pose  & Unseen-Pose    \\
 \hline
 CoKe-SHG & 94.0 & \textbf{84.0}\\
 SHG &   \textbf{94.2} & 73.6\\
 \hline
\end{tabular}
\caption{Robustness of keypoint detectors to unseen poses. For the \emph{car} category, we separate PASCAL3D+ training and testing set into 4 bins using azimuth annotations. We train CoKe on the front and back subsets, and test under both seen and unseen poses.}
\label{tab:experiment:unseenview}
\end{table}

\textbf{MPII.} We compare CoKe learned from the SHG backbone to a number of related works on MPII in Table \ref{tab:experiment:MPIIOri}. CoKe-SHG is again highly competitive and outperforms related work by a small but significant margin. Table \ref{tab:experiment:MPII} compares CoKe-SHG and SHG for both occluded and non-occluded keypoint detection on the MPII dataset. We can observe that CoKe achieves significantly higher results on occlusion scenarios compared with SHG. When compared to a recent and highly competitive baseline such as RSNs \cite{cai2020learning}, we can observe that RSN achieves a higher performance on the full dataset, due to its advanced architecture that has more than \textit{four times} more parameters compared to our model. However, we observe that the model suffers from a relatively higher performance decrease when the humans are occluded, hence it is not as robust as our model that uses a standard Stacked-Hourglass backbone. We also see the potential for the performance of our model to further increase with more advanced backbones. 
\begin{table}[h]

\centering
\begin{tabular}{l|l}
\hline
\multicolumn{2}{c}{\textbf{MPII}}    \\ \hline
RecurrentPose\cite{belagiannis2017recurrent}  & 88.1 \\ 
PoseMachines\cite{wei2016convolutional} & 88.5 \\ 
DeeperCut\cite{insafutdinov2016deepercut} & 88.5 \\ 
PartHeatmap\cite{bulat2016human}             & 89.7 \\ 
SHG\cite{newell2016stacked}                 & 90.9 \\ 
DualPathNetworks\cite{ning2017dual}         & 91.2 \\ 
PoseRegression\cite{bulat2016human}     & 91.2 \\ \hline 
CoKe-SHG            & \textbf{91.4} \\ \hline
\end{tabular}
\caption{Keypoint detection results on MPII compared with regression based algorithms. }
\label{tab:experiment:MPIIOri}
\end{table}

\begin{table}[h]

    \centering
    \begin{tabular}{l|cc}
        \hline
        \multicolumn{3}{c}{\textbf{MPII}}\\
      \hline
          & Full & Occluded \\
      \hline
        SHGs&90.1&84.3\\
        RSN \cite{cai2020learning}&92.7&86.9\\
        CoKe-SHG&91.4&86.7\\
      \hline
    \end{tabular} 
    \caption{Keypoint detection results on MPII of CoKe-SHG and SHG on both original and challenging scenarios.}
    \label{tab:experiment:MPII}
\end{table}
\begin{figure*}[!h]
\centering
\includegraphics[width=0.24\textwidth, height=0.14\textwidth]{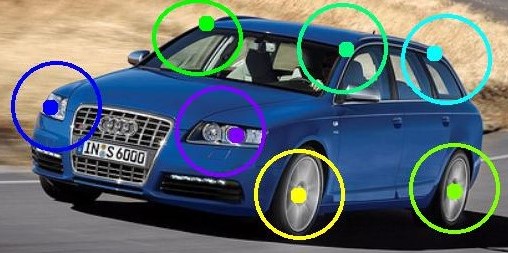}
\includegraphics[width=0.24\textwidth, height=0.14\textwidth]{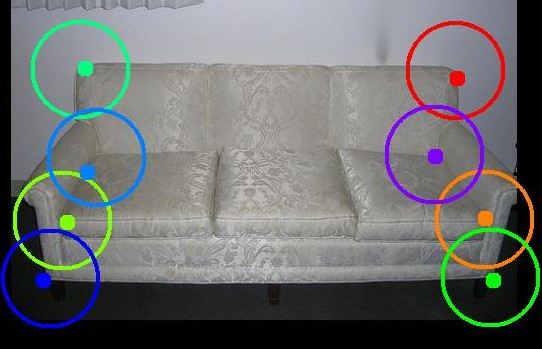}
\includegraphics[width=0.24\textwidth, height=0.14\textwidth]{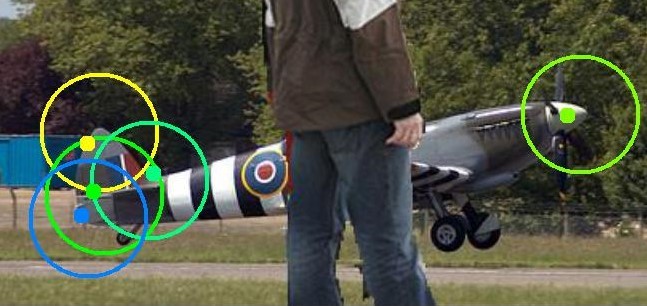}
\includegraphics[width=0.24\textwidth, height=0.14\textwidth]{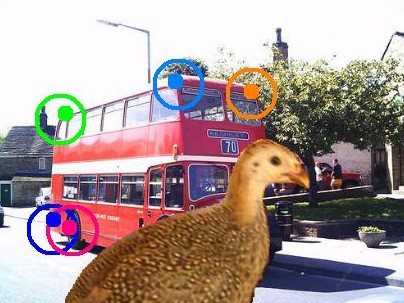}\\
    \includegraphics[width=0.24\textwidth, height=0.14\textwidth]{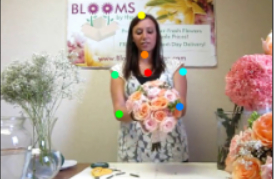}
    \includegraphics[width=0.24\textwidth, height=0.14\textwidth]{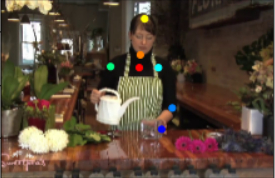}
    \includegraphics[width=0.24\textwidth, height=0.14\textwidth]{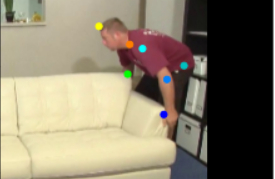}
    \includegraphics[width=0.24\textwidth, height=0.14\textwidth]{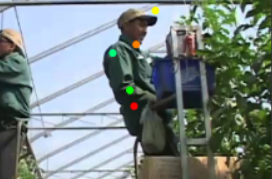}
\caption{Qualitative detection results under different levels of partial occlusion for artificially occluded objects from PASCAL3D+ and humans from MPII. The dots visualize the detection result of CoKe. The colored circles in indicate the ground-truth position within PCK=0.1, and indicate ground-truth position within PCKh=0.5 for MPII. Note how CoKe is very robust even under strong occlusion.}
\label{fig:occ_visualization}
\end{figure*}

\textbf{ObjectNet3D.} We compare SHG and a CoKe-SHG on ObjectNet3D in Table \ref{tab:experiment:objectNet3D}. CoKe-SHG outperforms SHG by a large margin on every category. Since ObjectNet3D dataset contains much more challenging scenarios, we claim that CoKe is more robust.

\begin{table}[h]

\begin{tabular}{l|ccccccc|c}
\hline
\multicolumn{6}{c}{\textbf{ObjectNet3D}}\\
 \hline
& coffee  & dryer   & kettle  & jar  & wash   \\
 \hline
 SHG &  31.0 &  35.1  & 32.2 &   41.9 &33.9    \\
 \hline
 CoKe-SHG &  \textbf{36.4} & \textbf{37.6}  &\textbf{37.8}&  \textbf{45.2}  &  \textbf{36.1}\\
 \hline
\end{tabular}
\begin{tabular}{l|ccccccc|c}
 \hline
& can  & calc  & eyegls      & guitar  & mean \\
 \hline
 SHG & 66.8  & 52.3   &  44.6      & 45.8  &  42.62\\
 \hline
 CoKe-SHG & \textbf{70.2}  & \textbf{57.7}   &  \textbf{51.3}  & \textbf{49.6} & \textbf{46.88}\\
 \hline
\end{tabular}
\caption{Keypoint Detection results on ObjectNet3D+~\cite{xiang2016objectnet3d}. 
}
\label{tab:experiment:objectNet3D}
\end{table}
In summary, we observe that CoKe is a general purpose framework that constantly achieves a very high performance across a wide range of backbone architectures and for a range of datasets with very different characteristics, while also being highly robust to occlusion.

\subsection{Qualitative Results}\label{ref:exp:qual}
\textbf{Visualization of Keypoint Detection Maps.} We visualize the part detection maps from CoKe-Res50 in Figure \ref{fig:featuremap}. Images are selected from the \emph{car} category in PASCAL3D+ which has complex changes in pose, illumination and object structure. Note how all keypoints are detected accurately despite the difficulty from false positives, occlusions, rare viewpoints, different domain, irregular appearance. We can observe that CoKe is robust in  these challenging scenarios.

\textbf{Visualization of Detection Results under Occlusion.} We visualize qualitative results in Figure \ref{fig:occ_visualization}.
Overall, the illustrations demonstrate the robustness of CoKe to partial occlusions. 
Any keypoints that are not in the vicinity of occluders are correctly detected and not affected by the occlusion. 
Furthermore, keypoints that are partially occluded (e.g. the wheel) can still be located robustly, although the detections tend to move away from the occluder. 
Importantly, we do not observe false positive detections at locally ambiguous keypoints.
This  demonstrates that CoKe leverages the receptive field to disambiguate keypoints, while still being able to localize individual keypoints accurately. 
\textbf{Inference Time and Memory Consumption.} During inference, CoKe-Res50 (params: 23M, acc: 77\%) takes 0.01s per image, while SHGs (params: 25M, acc: 68\%) needs 0.06s. For the memory consumption, CoKe-Res50 needs 715MB, SHGs 786MB when batch size equals to 1. 
CoKe has an advantage of the inference time while maintaining the competence of the memory consumption.

\subsection{Ablation Study}
\label{sec:experiment:ablation}
\textbf{Comparison with Other Losses.}
In Table \ref{tab:experiment:p3dlosses}, we ablate the effect of the contrastive learning compared to other common losses on different backbone architectures.
Specifically, we compare to a regression loss as used in SHGs and a supervised classification loss. 
The classification baseline enables us to compare the effect of the contrastive learning with a cross-entropy loss. 
We implement keypoint detection as a classification problem by using the features at the ground truth keypoint locations in the training data as instances of different classes to train the backbone with the cross-entropy objective. 
During testing, we use the average representations from the training set as part detector, similar as in the CoKe pipeline. 
The main difference to the CoKe training is hence that the contrastive loss and the clutter bank are not used. 
From the results, we observe that our representation learning formulation of keypoint detection constantly outperforms other losses by a large margin. 
\begin{table}

    \centering
    \begin{tabular}{l|ccc}
      \hline
      \multicolumn{4}{c}{\textbf{PASCAL3D+}} \\
      \hline
      Backbone & Reg & Class & CoKe \\
      \hline
      SHGs & 68.0 & 67.8  & \textbf{78.3} \\
      Res50 & 68.9 & 69.3  & \textbf{77.0}\\
      ResUnet & 69.7 & 70.2  & \textbf{80.3}\\
      \hline
    \end{tabular}
    \caption{Ablation study of CoKe training strategy in comparison of standard regression or classification training loss.
    }
    \label{tab:experiment:p3dlosses}
\end{table}

\begin{table}

\centering
\begin{tabular}{lcccc}
\hline
\multicolumn{1}{l|}{Occlusion Level}                                                                    & Lv.0                                      & Lv.1                                      & Lv.2                                      & Lv.3                                      \\ \hline
\multicolumn{1}{l|}{No clutter}                                                                         & 79.3                                      & 75.4                                      & 71.8                                      & 65.8                                      \\ 
\multicolumn{1}{l|}{Image-specific clutter}                                                             & 92.8                                      & 82.7                                      & 76.5                                      & 69.2                                      \\ \hline
\multicolumn{1}{l|}{Clutter Bank (64 groups)}                                                           & 93.0                                      & 83.6                                      & \textbf{80.1}                                      & \textbf{73.3}                                      \\ 
\multicolumn{1}{l|}{Clutter Bank (256 groups)}                                                          & 94.3                                      & 84.3                                      & 77.7                                      & 71.0                                      \\ 
\multicolumn{1}{l|}{Clutter Bank (1024 groups)}                                                         & \textbf{95.5}                                      & \textbf{85.9}                                      & 79.0                                      & 70.6                                      \\ \hline
\multicolumn{1}{l|}{\begin{tabular}[c]{@{}l@{}}Clutter Bank w/o  \\ clutter sampling loss\end{tabular}} & 94.2                                      & 83.1                                      & 76.8                                      & 68.0                                      \\ \hline
\end{tabular}
\caption{Ablation study on PASCAL3D+ with different settings: no clutter features, image-specific clutter features (using 20 features from the same image as clutter examples), our proposed Clutter Bank with different number of groups (each group contains 20 features) and deactivating clutter loss. Note the benefit of using clutter features in general, and in particular using a large clutter bank, as well as the importance of the clutter sampling loss.}
\label{tab:experiment:ablation}
\end{table}
\textbf{Clutter Bank Mechanism}
In Table \ref{tab:experiment:ablation}, we study the influence of the clutter features and the clutter sampling loss on the contrastive learning result. In particular, the table shows the keypoint detection results for CoKe-Res-UNet on the car category of the PASCAL3D+ dataset.
We observe that the performance decreases significantly when no clutter features are used during training. 
An extension of this basic setup is to use image-specific clutter features but without maintaining the features in a Clutter Bank.
In particular, we select the clutter features from the same image from which the keypoint features are sampled using the same hard negative sampling mechanism as in our standard setup. From the results we observe that using the image-specific clutter features increases the performance significantly.
However, the best performance is achieved using our proposed Clutter Bank mechanism. 
In particular, the results show a general trend that the more features we store in the bank, the higher the performance becomes. 
Notably, lower occlusion scenarios benefit from a larger clutter bank while for stronger occlusion a smaller clutter bank is more beneficial.
Finally, our ablation shows that explicitly regularizing the clutter to be distinct from the Keypoint Bank using the clutter sampling loss is highly beneficial.\\

\textbf{Cumulative Moving Average Update}
We also study the importance of the cumulative moving average update. Here we provide another possible method: average approximation. In particular, we calculate an average over the whole dataset for $\{\mathbf{\theta}\}$ each 10 epochs. We report the result using CoKe-Res-UNet on the car category of the PASCAL3D+ dataset. 
We observe a deduction of keypoint detection accuracy as L0: $94.2 \rightarrow 88.7$, L1: $83.1 \rightarrow 80.0$, L0: $76.8 \rightarrow 75.0$, L0: $68.0 \rightarrow 68.9$.
Conclusively, the best performance is achieved using our proposed cumulative moving average update mechanism. 

\section{Conclusion}
In this paper, we study keypoint detection from the perspective of contrastive learning.
Our contrastive keypoint learning framework (CoKe) makes several efficient approximations to enable the contrastive representation learning for keypoint detection: (i) A clutter bank to approximate non-keypoint features; (ii) a keypoint bank that stores prototypical representations of keypoints, to approximate the within-keypoint distance; and (iii) a cumulative moving average update to learn the keypoint prototypes while training the feature extractor.
Our experiments on several diverse datasets (PASCAL3D+, MPII, ObjectNet3D) show that CoKe is very general and performs well for rigid as well as articulated objects. Compared to related work, CoKe achieves higher performance, while also being much more robust to partial occlusion and unseen object poses.

\noindent \textbf{Acknowledgements.} AK acknowledges support via his Emmy Noether Research Group funded by the German Science Foundation (DFG) under Grant No. 468670075.

{\small
\bibliographystyle{ieee_fullname}
\bibliography{coke}
}

\end{document}